# Constellation Design of Remote Sensing Small Satellite for Infrastructure Monitoring in India

Roshan Sah, Raunak Srivastava, Kaushik Das
TCS Research
Bangalore, India; +919933422226
sah.roshan@tcs.com

**ABSTRACT**

A constellation of remote sensing small satellite system has been developed for infrastructure monitoring in India by using Synthetic Aperture Radar (SAR) Payload. The low earth orbit (LEO) constellation of the small satellites is designed in a way, which can cover the entire footprint of India. Since India lies a little above the equatorial region, the orbital parameters are adjusted in a way that inclination of 36 degrees and RAAN varies from 70-130 degrees at a height of 600 km has been considered. A total number of 4 orbital planes are designed in which each orbital plane consisting 3 small satellites with 120-degrees true anomaly separation. Each satellite is capable of taking multiple look images with the minimum resolution of 1 meter per pixel and swath width of 10 km approx. The multiple look images captured by the SAR payload help in continuous infrastructure monitoring of our interested footprint area in India. To support the mission, each small satellite is supplied with earth sensors, sun sensors, GPS to accurately determine its position and attitude, and Control Moment Gyro (CMGs) which is capable of high slew rate maneuver with precise pointing at minimum power utilization. Further, each small satellite is equipped with a communication payload that uses X-band and VHF antenna, whereas the TT&C will use a high data-rate S-band transmitter. The satellite requires a powerful set of batteries to operate along with an origami-designed solar panel with the implementation of GaN-FETs to improve the performance and efficiency of solar power conversion. The paper presents only a coverage metrics analysis method of our designed constellation for our India footprint by considering the important metrics like revisit time, response time, and coverage efficiency. The data processing for the captured images is not presented here. The result shows that the average revisits time for our constellation ranges from about 15- 35 min which is less than an hour and the average response time for this iteratively designed constellation ranges from about 25-120 min along with hundred percent coverage efficiency most of the time. Finally, it was concluded that each satellite has 70kg of total mass and costs around $ 0.75M to develop.

## INTRODUCTION

In the twentieth first century, small satellite technology has played an impactful role in the development of space disruptive technology. The technology trends have made an important advancement in small satellites within the range of 1-500 kg weight. The small satellite are mainly classified as Femto (<0.1kg), Pico (0.1-1 kg), Nano (1-10 kg), Micro (10-100 kg) and Mini (100-500 kg) satellite. Numerous space technology companies from different countries have been started working related to different payload applications in small satellite setup which will be changing the space economy exponentially. It has been found that the global small satellite market will grow from 2.8 to 7.8 billion dollars by the end of 2025 as updated by the small satellite market report [1]. Mostly, Nano-satellite and micro-satellites have been used for testing, experimenting, and exploring new space hardware or software along with the new ideas at a minimum expense of money. In the nineteenth century, single spacecraft was enough to perform multiple missions in space, but recently, it can be seen that for some of the advanced space applications, the single spacecraft is not enough to fulfill the desired mission in space. To fulfill these requirements, a series of satellites working together can be used in the same or different orbit, known as "Satellite Constellation."

In the recent decade, the satellite constellation idea has evolved broadly, with huge satellites operating in space to fulfill the required application demand. It is used to permanently provide communication facilities and global coverage at anytime, anywhere on the earth. There are lots of operating satellite constellation that are used for an application like navigation and positioning constellation (e.g., GPS, NAVIC, GLONASS, Galileo, BeiDou, QZSS, Etc.), communication (Global Xpress, Globalstar [2], European Aviation Network, Broadband Global Area Network, O3b, Eutelsat, Thuraya, Etc.), earth observation satellite constellation (RADARSAT Constellation, Planet Labs, Rapid Eyes, Disaster Monitoring Constellation, Spire, Etc.), remote sensing, Etc. The access to space is continuously increasing due to the design and technology miniaturization of large satellites to small satellites and an increase in small satellites' launch rate from Femto to mini size [3]. Moreover, their constellation is getting tremendous attention these days for sustainable business purposes [4]. The upcoming decade is IoT, edge computing, weather science, disaster monitoring, and safety/security, which will have the most significant potential and dependency in the satellite constellation development for the business purpose [5]. More than one hundred companies had made a proposal for a satellite constellation in different with different approaches in recent years.

Nowadays, satellite constellation has become a powerful and effective tool for earth observation and remote sensing applications. Remote sensing is one of the most famous applications in satellite technology for the development of technology like infrastructure, disaster, weather, biodiversity, forestry, surface change, and agriculture monitoring purpose. Previously, most of the constellations used to be at Sun Synchronous Orbit (SSO) or High Earth Orbit (HEO) for remote sensing and earth observation with larger satellites. But in recent trends, most of the satellite constellation is often deployed at Low Earth Orbit (LEO), as the single satellite only covers certain small areas of the earth that orbit at higher angular velocity to maintain its orbit. Therefore, numerous LEO satellites are required to provide continuous permanent global coverage. In recent years, satellite constellations are continuously evolving as the industry market's business solution for providing global, regional coverage, and space research. The regional constellation satellite-like Indian, Regional Navigation Satellite System (IRNSS) [6], and Quasi-Zenith satellite [7] systems are continuously involved in providing the local region coverage to people. As regional coverage mostly focuses on local area coverage, a small satellite system must provide the same performance per metric area compared to global constellation coverage. Regional constellation performance leads to a reduced launch and design cost system, which is directly dependent on the number of satellites used [8].

In this paper, we will present our work in each section. The section consists of constellation design, preliminary sizing of the satellites, coverage quality measurement, results, and discussion of our simulated case and overall conclusion. The constellation design represents the iterative design procedure of selection of our orbital parameters and orbital plane, which will be capable of making overall coverage of our area of interest footprint. Whereas the preliminary sizing section represents the distribution of the mass and cost budget of each sub-system of the satellites and their selection of the optimal components required while making a full working satellite. This section also represents our SAR payload specification which is required to take multiple images while passing over India. The coverage quality measurement section explains the method of finding the different coverage metrics and mathematical models involved in it along with the importance of the coverage performance of our designed constellation. The image and data processing for the captured images is not presented in this paper as it will be done by other groups. The result and discussion section show the important results of some coverage metrics parameters which are required to evaluate the coverage performance of our constellation. Finally, all the section is concluded in the conclusion section.

**CONSTELLATION DESIGN**

The traditional way of designing the satellite constellation mainly focuses on minimizing the number of satellites to provide continuous broad area coverage. The design process mainly involves determining the typical characteristics of orbit and constellation patterns to generate the optimal constellation for complex coverage requirements. The classical constellation design method, like Rosette constellation [9], Walker constellation [10], and street of coverage [11], and tetrahedron elliptical constellation [12] which provides us the geometric design approach for uniform symmetric constellation pattern. The symmetric pattern provides a solid foundation for a fully optimized design space analysis to get an analytical solution due to the finite variability approach [13]. As the traditional constellation mostly assumes symmetric patterns and fully optimized common orbital characteristics, the spatial-varying and time-varying coverage requirements are completed missed in large, designed space.

In this paper, we had presented LEO small satellite constellation design in an iterative way, which will be capable of capturing the entire footprint coverage of the India location in 24-hour duration. Each small satellite consists of a SAR payload system which enables to capture of the footprint at each revisit to the India location. The iterative design of the satellite constellation and orbital parameter were run on the AGI STK [14] tools by considering the SGP4 [15] orbital propagator to minimize the positional and attitude discrepancy. The selected orbital elements and the number of the satellite in each orbital plane are shown in Table 1.

Since India lies a little above the equatorial region; the orbital parameters are adjusted with higher access time and coverage with an inclination of 36° and altitude of 600 km. To make full coverage of the India footprint in 24 hours, we had considered a total of 12 numbers of the small satellites with 4 different inclined orbital planes.

**Table 1. Small Satellite Constellation Orbital Parameters.** (Whereas G1, G2, G3, and G4 are the individual orbital planes.)

|    | Parameters           | G1     | G2     | G3     | G4     |
|----|----------------------|--------|--------|--------|--------|
| 1. | Height               | 600 km | 600 km | 600 km | 600 km |
| 2. | Inclination (i)      | 36°    | 36°    | 36°    | 36°    |
| 3. | RAAN                 | 70°    | 90°    | 110°   | 130°   |
| 4. | No. of Satellites    | 3      | 3      | 3      | 3      |
| 5. | True Anomaly separation | 120° | 120°   | 120°   | 120°   |

Each orbital plane consists of the 3 number of the satellite which was separated by 120 degrees of the true anomaly in an orbit. The RAAN varies from 70-130 degrees across the plane with a difference of 20 degrees between the orbital planes. By considering mentioned orbital plane, each satellite will have at least 3 accesses to the India location.

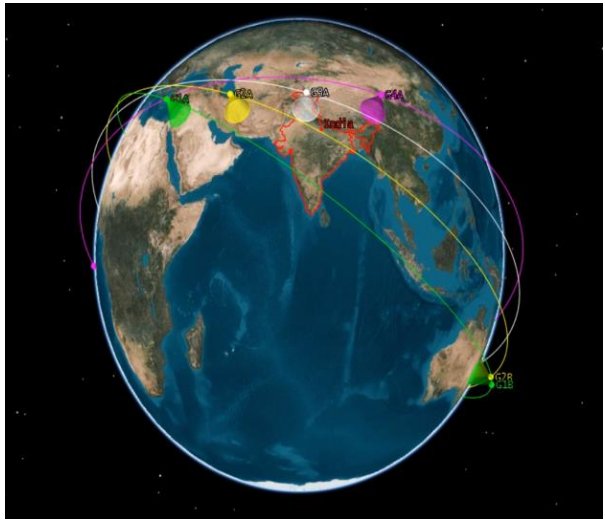

**Figure 1: Small satellite constellation design visualization in 3D space.**

The schematic diagram of a small satellite constellation having 4 orbital planes with 3 satellites at each plane along with the India footprint is shown in figure 1 in 3D visualization which was generated by the means of STK tools. Figure 2 represents the ground track of the satellite to the India location along with the access coverage which is marked by the red color.

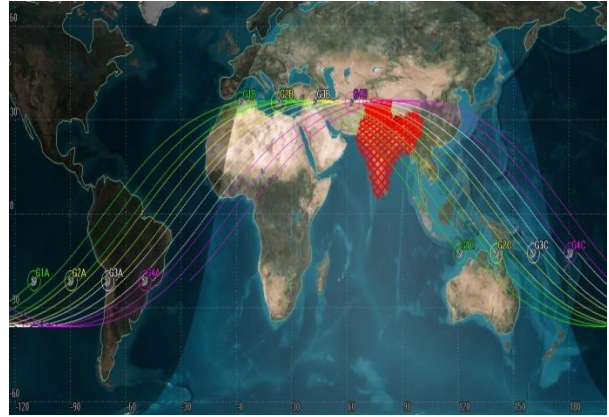

**Figure 2: Small satellite constellation coverage access in India footprint.**

### PRELIMINARY SIZING OF THE SATELLITES

The design of the satellite always starts with the conceptual and type of payload is used for the mission. Once the payload is known, one can go for the selection of the components of each sub-system based on the availability in the market. The selected component of each sub-system is shown in the form of the UML diagram. The Unified Modeling Language (UML) diagram is kind of a flow diagram, which represents the input and output of the system. The UML diagram of our selected components for each sub-system is shown in figure 3.

From figure 3, we can see that the entire optimal component required for our satellites is represented by each sub-system box which will act as input to our INFRASAT satellite system. As the output, we will get different parameters like mass budget, power budget, point accuracy, optimal attitude requirement, etc., which were categorized as each sub-system requirement. These mentioned parameters from the diagram are required while designing our whole set of a satellite. Each small satellite is equipped with a communication payload that uses X-band and VHF antenna, whereas the TT&C will use a high data-rate S-band transmitter. The satellite requires a powerful set of batteries to operate along with an origami-designed solar panel with the implementation of GaN-FETs to improve the performance and efficiency of solar power conversion. The mass and cost budget estimations are the crucial part of the design which helps us to design each component by the method of compact engineering to the satellite body. The mass determination of each sub-system is done from the existing literature [16] and the cad design mass method. We had used both method and approximate mass budget were found based on the availability of the sub-system component in the market.

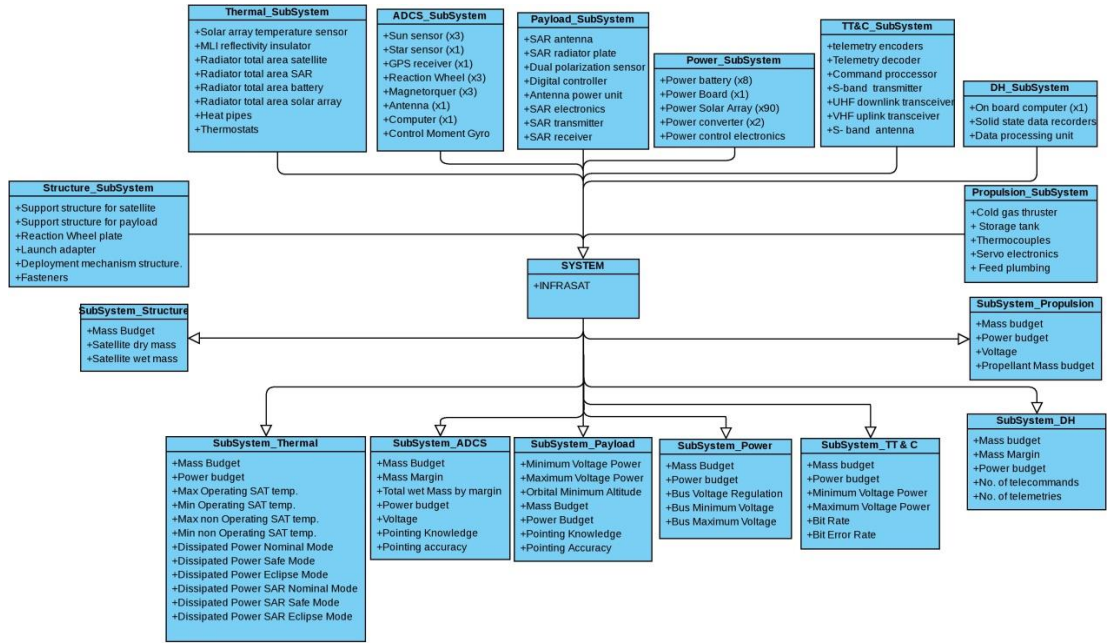

**Figure 3: UML diagram of InfraSat essential sub-systems component with essential parameters.**

**Table 2: InfraSat's Mass and the Cost budget.**

| Sub System | Mass [Kg] (% of total) | Cost Budget (Lakhs) (% of total) |
|---|---|---|
| Structures | 10.55 (15.10%) | 53 (9.62%) |
| ADCS | 3.21 (4.60%) | 78 (14.15%) |
| Thermal | 2.92 (4.17%) | 20 (3.63%) |
| Power | 16.57 (23.70%) | 95 (17.24%) |
| Communication | 2.60 (3.71%) | 61 (11.07%) |
| C &DH | 2.25 (3.22%) | 54 (9.8%) |
| SAR Payload | 25.50 (36.48%) | 130 (23.60%) |
| Propulsion | 5.50 (7.87%) | 20 (3.63%) |
| Sub-total Integration | 0.8 (1.14%) | 10 (1.81%) |
| Miscellaneous | - | 30 (5.44%) |
| Total | 69.90 (100%) | 551 Lakh (100%) |
| Target | 70-72 kg | |
| Margin | 0.10-2.1 kg | |

The mass budget for our SAR payload satellites is shown in Table 2.

The total estimated mass budget was 70-72 kg according to our calculation, but by cad design mass calculation, we got approximately 69.9 kg. The mass values include the contingencies. The payload sub-system consists of a maximum mass of about 25.5 kg in the whole satellite system along with the power system which has a mass of about 16.57kg. And we will be adding a micro-propulsion system about of 5.5 kg weight for maneuvers purpose.

The Cost of the SAR small satellite is determined by the Small Satellite Cost Model 2019 (SSCM19) procedure. The Cost of the satellite mostly depends upon the mass budget of the subsystem. SSCM19 model estimates the costs of the satellite weighing less than 1000 kg by using Cost Estimating Relations (CERs) procedure which was derived from technical and actual cost parameters [17]. The cost budgets for our SAR payload small satellite are shown in Table 2. The cost budget values also include the contingencies. It can be seen that the total cost of each remote sensing satellite is about 551 Lakh INR which is approximate 0.75M $. From, overall cost, the SAR payload and power sub-system costing a higher amount for the manufacture than other subsystems. We will be adding the micro-propulsion system for our satellite which will help us to execute different maneuvers and to de-orbit the satellite to the lower orbit after the end of life (EOL) of a satellite.

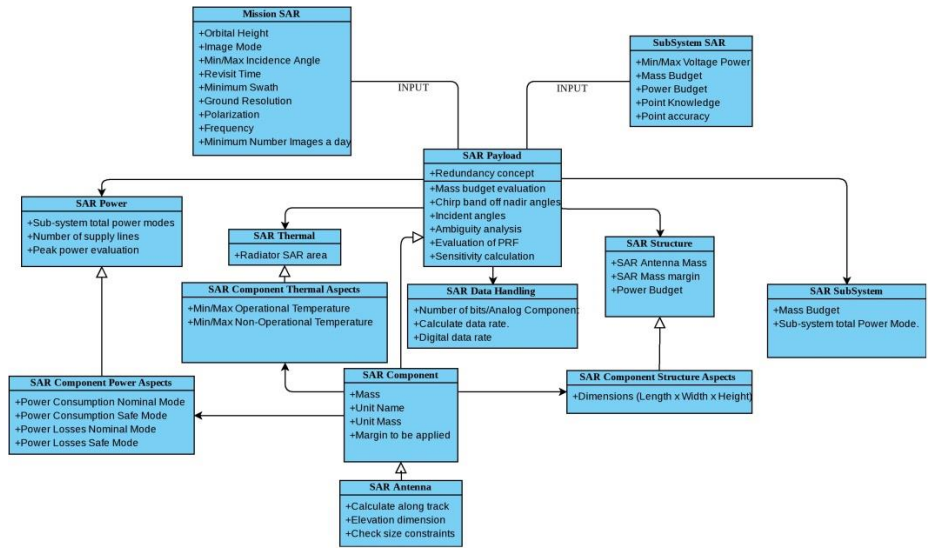

**Figure 4: UML diagram of InfraSat SAR payload essential parameters.**

The UML diagram of SAR payload considered parameters while designing is shown in figure 4. Based on our mission objective, the design for the SAR payload is considered. The SAR payload design depends upon all the subsystem requirements. The diagram represents the optimal parameters required by each subsystem and component while making operation SAR in the LEO space environment. It has shown structural, thermal, data handling, and power parameters or aspect for our SAR payload. The arrow represents the flow characteristics required while making our SAR payload to the different subsystems of the satellite. This diagram has also shown the parameter like elevation dimension, size constraint check, and along-track calculation to be considered while designing the SAR mesh reflector antenna. The antenna becomes the most important component in the SAR payload which the main function is to receive and transmit data to the ground station installed in our footprint.

The specification of our InfraSat payload is shown in Table 3. Our SAR payload consists of a mesh reflector antenna having an optical sensor with an x-band sensor having a frequency band of 200 MHz which will be operating on spotlight mode. The spotlight mode provides high-resolution images of our target footprint by using parabolic antennas, which can be rotated by mechanical means. Each satellite will be capable of taking multiple look images by using the spotlight mode with the minimum resolution of 1 meter per pixel and swath width of 10 km approx. The ground sample distance (GSD) for it is less than 0.5/2 m. The satellite will be taking multiple images from an orbit about of 3 min to our footprint area. The taken images will be transmitted to the ground station by using the x-band frequency at the rate of 500 Mbps. Each satellite will be having a design life of 4 years.

**Table 3: Specification of InfraSat SAR payload.**

| Item | Specifications |
| --- | --- |
| Sensor | Optical |
| Sensor frequency | X-band |
| Antenna type | Mesh reflector |
| Observation Mode | Spotlight mode |
| GSD | <0.5/2 m (multi from 600 km) |
| Polarization | Linear |
| Resolution | ~1m |
| Swath | 10 km (spotlight mode) |
| Data transmission | X-band, 500Mbps, 10QAM |
| Bandwidth | 200 MHz |
| Imaging time per orbit | 3 min |
| Altitude | 600 km (Nominal) |
| Design life | 4 years |

**COVERAGE QUALITY MEASUREMENT**

To calculate the satellite coverage towards India's footprint, we just need to understand the different satellite coverage characteristics and its performance.

The coverage characteristic consists of coverage quality or metrics as the preliminary requirement which will address the ground track across our area of interest footprint. The coverage metrics help us to investigate the regional or global coverage provided by one or more satellites by considering all the access times of the satellites.

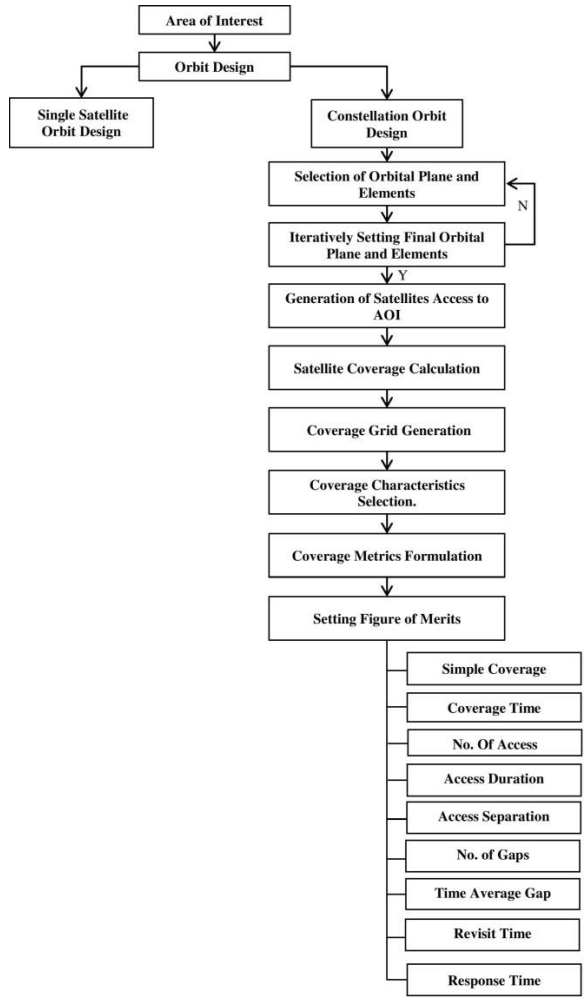

**Figure 5: Block diagram for orbital design and coverage metrics calculation for India footprint.**

The schematic block diagram to find the coverage characteristics and parameters for our India location AOI is shown in figure 5. Once the constellation, orbital plane, and elements are designed, the satellites undergo the simulation phase. The simulation of the constellation was run for one day and the satellite access parametric dataset was generated for our given AOI. These datasets will help us to calculate the satellite coverage parameters. Before finding it, we need to generate the grid/mesh between the latitude and longitudinal direction of our AOI. The grid granularity of India's location is discretized /meshed at 0.5° across the longitudinal and latitudinal direction. The coverage grid uses a set of points for the coverage sampling within our AOI. The schematic diagram of the discretized grid for the India location is shown in figure 6.

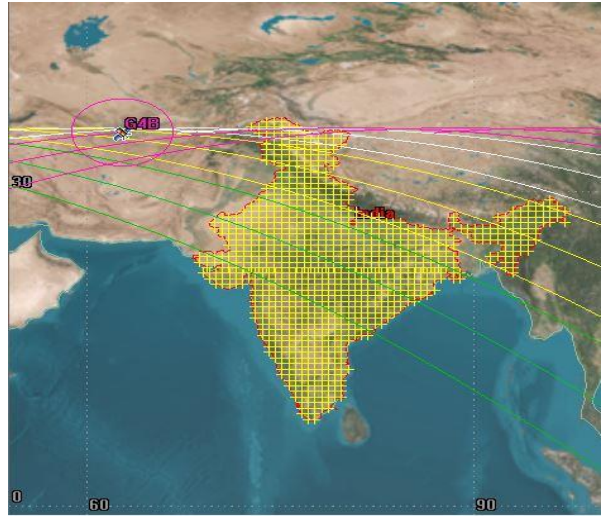

**Figure 6: Coverage grid generation of India footprint with 0.5° discretization along lat-longs.**

Once the coverage grid is generated then the coverage characteristics are selected which helps us to formulate the coverage metric via the figure of the merits. Based on the access time from each satellite to each generated coverage grid point, the coverage iteratively computes the figure of merits satisfaction.

The figure of the merits (FOM) consists of the following coverage quality metrics [18, 19];

- Simple Coverage: It is defined as the coverage measured by the satellite at a point.
- Coverage Time: It is the time taken by satellites to cover the given target grid points in a given time. It can be measured in maximum, average, and total coverage time.
  It can be expressed as;

$$T_{c,total} = \sum_{i=1}^{N} T_c(i) \qquad (1)$$

$$T_{c,mean} = \frac{T_{c,total}}{N} \qquad (2)$$

$$T_{c,max} = \max\{T_c(i)\} \qquad (3)$$

Where, $T_{c,total}$, $T_{c,mean}$, $T_{c,max}$ and $T_c$ represents the total, mean, maximum, and single coverage time. N and i represent the total number of the coverage duration in i-th time. If the maximum coverage time is larger, there will be enough time to capture an image for infrastructure monitoring.

- No. of Access: It generally counts the coverage access by the set of satellites independently towards a point.
- Access Duration: It is the time interval taken by the satellites between the individual coverage accesses.
- Access Separation: It generally measures the coverage from the multiple satellites to a point within a defined time limit.
- No. of Gaps: It counts the gap number formed during the coverage to the points.
- Time Average Gap: It is the average time taken by no. of the gaps formed during coverage of given grid points.
- Revisit time: It is the time interval at which the satellite continuously covers the same grid point. The revisit time is represented in maximum, minimum, and average format. But the average revisit time is mostly used for the analysis purpose and is the important performance parameter in coverage metrics. It can be expressed as;

$$T_{R,av} = \frac{1}{N} T_c(i) \quad (4)$$

Where, $T_{R,av}$ represents the average revisit time duration of the satellite.

- Response time: It is the difference in the time at which coverage at a given grid point is obtained and coverage request is made.

It is computed as

$$T_{Re,av} = \frac{T_{Av,gap}}{2} \quad (5)$$

Whereas, $T_{Av,gap}$ represents the time average gap.

Out of all the coverage metrics, the coverage time, revisit time, and response time are the major requirements for most of the infrastructure monitoring remote sensing satellites. These metrics help in determining the coverage characteristic performance of the constellation satellites. The satellite constellation can get the benefit of the higher coverage time and short revisit time to the given area of interest (AOI).

## RESULTS AND DISCUSSION

For our infrastructure monitoring, the coverage metrics are the most important aspect of our constellation design for SAR payload. Out of all the coverage metrics, the revisit time, response time, and percentage coverage are the most important metrics for our analysis. So, we had found these metrics based on our constellation design by considering the SAR payload.

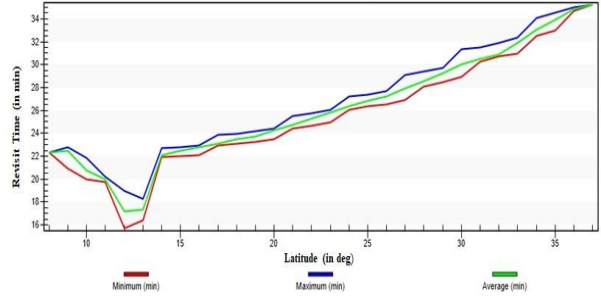

**Figure 7: Graphical representation of revisit time (in min) across latitude (in degree).**

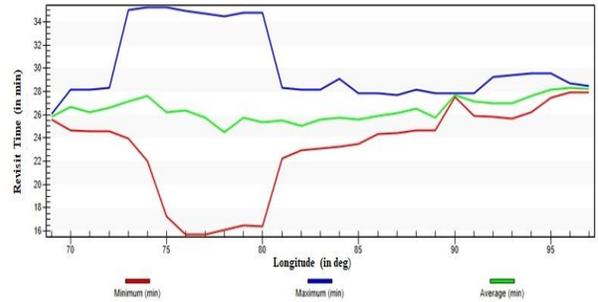

**Figure 8: Graphical representation of revisit time (in min) across longitude (In degree).**

The schematic diagrams of revisit time along latitude and longitudinal location of India footprint are shown in Figures 7 and 8. From the graphs, we can see that the revisit time for our footprint by considering all the satellites operating in designed constellation ranges from 15- 35 minutes in any location of the India. The graphs also indicate the minimum, maximum and average values of the revisit time in the allocated footprint.

The schematic diagram of the average revisit time contour for our India footprint can be shown in figure 9 by using the data set of average revisit time values along with latitude and longitude of footprints. The different color code represents average revisit time values of a particular location. And the boundary is encircled by a red color mark. This contour helps us to find the average time is required by satellites to reach again to the same point in an area of India. By knowing the average revisit time, we will be capable of finding when the satellite will be in our interested city and places of India. So, the operation of the SAR payload in spotlight mode can be done and multiple look images can be taken for allocated infrastructure like building, oil pipes, canals, agriculture lands, roads, traffic monitoring, etc. These images obtained of allocated areas can be used for the monitoring and surveillance of that area.

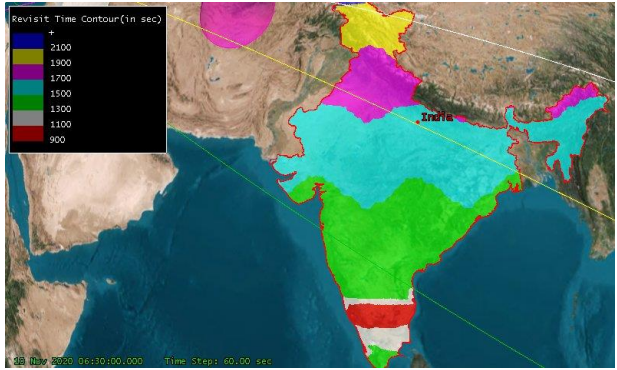

**Figure 9:- Average revisit time contour representation across the footprint.**

Similarly, we had found the response time required for the smooth operation of the constellation of the satellite in the allocated footprint. The schematic diagrams of response time along the latitude and longitude location of India are shown in Figures 10 and 11. The response time was found in minimum, maximum, and average values in an allocated area of the footprint.

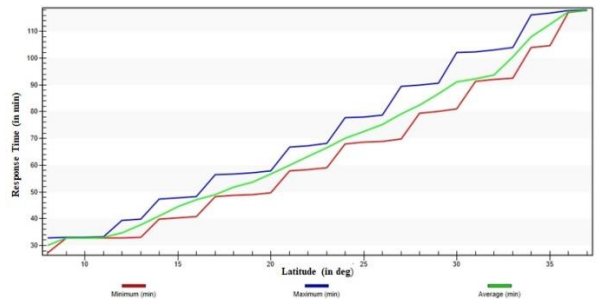

**Figure 10: Graphical representation of response time (in min) across latitude (in degree).**

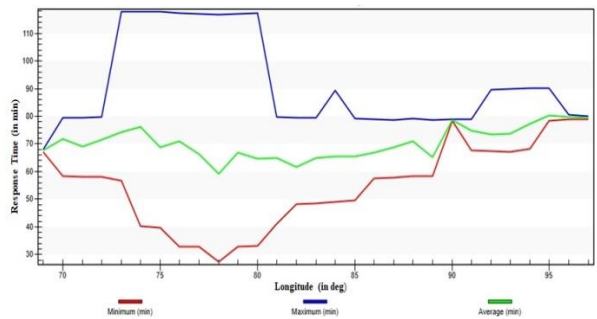

**Figure 11: Graphical representation of response time (in min) across Longitude (in degree).**

From both graphs, it can be seen that the response time value ranges from 25-120 min in any location of our footprint.

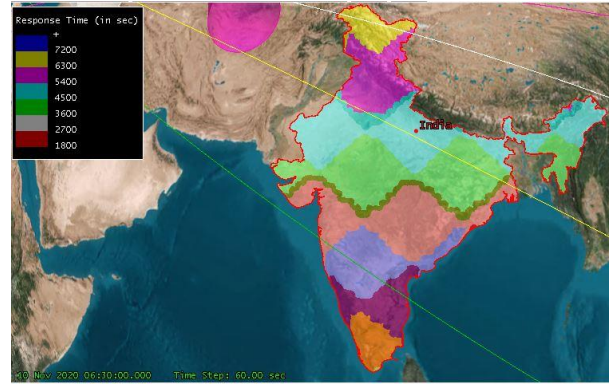

**Figure 12: Average response time contour representation across the footprint.**

The schematic diagram of the average response time contour for India's footprint is shown in figure 12. From the figure, it can be seen that the different color codes can help us to find in which area, the response time required is more or less. These average values help to find at which time, we need to make the coverage request so the full coverage at the given grid location can be obtained. These metrics helps the payload sub-system to get ready for its operation.

The graphical representation of the coverage percentage along the simulation time is shown in figure 13. We can see that the coverage percentage is varying with respect to time and at a certain time; there is a gap which means we don't have satellite access during that phase of time. But most of the time, the efficiency of the coverage is about a hundred percent.

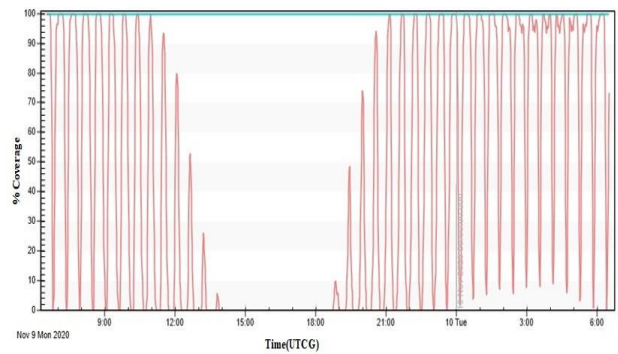

**Figure 13: Graphical representation of % coverage across the simulation time (in UTCG).**

Overall, we can see that the revisit time, response time, and coverage efficiency in the coverage metrics play a primary role in the coverage performance of our satellites. These metrics help us to predict when the satellite will be coming to our interested footprint and when we can have our payload operation during the mission. These parameters can act as guidelines for making our different subsystems in a highly accurate manner.

**CONCLUSION AND FUTURE WORKS**

A constellation of the small satellites for infrastructure monitoring has been designed by using the SAR as a payload. This constellation will be developed at LEO orbit which will cover an entire footprint of India location. The preliminary design of the satellite is done based on an available product in the market and on essential parameters required for our operation. In this paper, we had presented our work in each section. The section of constellation design has represented the iterative design procedure of selection of our orbital parameters and orbital plane. And it was found that a total of 4 orbital planes were designed. Each orbital plane contains the 3 satellites and a total of 12 satellites will be operating. The orbital parameters are adjusted in a way that inclination of 36 degrees and RAAN varies from 70-130 degrees at a height of 600 km has been considered, which will be capable of making overall coverage of our India's footprint.

Whereas the preliminary sizing section has represented the distribution of the mass and cost budget of each sub-system of the small satellites and our SAR payload specification which is required to take multiple images while passing over India. Each InfraSat estimates a cost budget of about $ 0.75 M with an approximate of 70 kg mass budget. It will have a life span of 4 years and make continuous coverage across India. Each satellite is equipped with a sun sensor, GPS receiver, control moment gyros which will help to determine the satellite attitude and position accurately at minimum power requirement. The SAR payload will operate in the spotlight mode with a minimum resolution of one meter per pixel at a swath of 10km radius. The SAR data processing of the captured image are not presented in this paper. We have only shown the coverage metrics calculation and the results of the designed constellation.

The coverage quality measurement section had described coverage metrics method and mathematical models involved in it along with the importance of the coverage performance. The coverage quality metrics were determined and essential metrics were calculated for the allocated footprint. Out of all the coverage metrics, the revisit time, response time, and coverage efficiency are the primary requirement to determine the coverage performance of the constellation. For our footprint, we found that the average revisits time for our constellation ranges from about 15- 35 min which is less than an hour. And the average response time for this iteratively designed constellation ranges about 25-120 min. Similarly, for the same constellation, we mostly got a hundred percent coverage efficiency most of the time.

For future work, these preliminary datasets and calculations will be used for a more detailed design of our constellation satellite. Based on the market requirement, the potential investigation and up gradation in the SAR sub-system will be done with alternative observation mode in consideration. Moreover, the operation of the SAR payload will cause high consumption of power; a market study will be done based on higher power availably with the least mass budget. Lastly, the other coverage metrics will be studied and analyzed for our designed constellation.